\documentclass[10pt,twocolumn,letterpaper]{article}

\usepackage{wacv}              
\usepackage[accsupp]{axessibility}

\usepackage{graphicx}
\usepackage{amsmath}
\usepackage{amssymb}
\usepackage{booktabs}

\usepackage{url}
\usepackage{amssymb}
\usepackage[utf8]{inputenc} 
\usepackage[T1]{fontenc}    
\usepackage{url}            
\usepackage{booktabs}       
\usepackage{amsfonts}       
\usepackage{nicefrac}       
\usepackage{microtype}      
\usepackage{bm}
\usepackage{enumitem}
\usepackage{graphicx}
\usepackage{amsmath}

\usepackage{caption}
\usepackage{algorithm}
\usepackage{algorithmic}
\usepackage[dvipsnames]{xcolor}
\usepackage{float}
\DeclareMathOperator*{\argmin}{arg\,min}
\DeclareMathOperator*{\argmax}{arg\,max}
\usepackage{array}
\usepackage{multirow}
\usepackage{wrapfig,lipsum,booktabs}

\usepackage[pagebackref,breaklinks,colorlinks]{hyperref}

\usepackage[capitalize]{cleveref}
\crefname{section}{Sec.}{Secs.}
\Crefname{section}{Section}{Sections}
\Crefname{table}{Table}{Tables}
\crefname{table}{Tab.}{Tabs.}

\def\wacvPaperID{781} 
\def\confName{WACV}
\def\confYear{2024}

\begin{document}

\title{On the Quantification of Image Reconstruction Uncertainty without Training Data}

\author{Sirui Bi  \\
Walmart Global Tech\\
{\tt\small siruijhu@gmail.com}
\and 
Victor Fung  \\
Georgia Institute of Technology\\
{\tt\small victorfung@gatech.edu}
\and 
Jiaxin Zhang\thanks{Corresponding author}\\
Intuit AI Research\\
{\tt\small jxzhangai@gmail.com}
\and
}
\maketitle

\begin{abstract}
Computational imaging plays a pivotal role in determining hidden information from sparse measurements. A robust inverse solver is crucial to fully characterize the uncertainty induced by these measurements, as it allows for the estimation of the complete posterior of unrecoverable targets. This, in turn, facilitates a probabilistic interpretation of observational data for decision-making. In this study, we propose a deep variational framework that leverages a deep generative model to learn an approximate posterior distribution to effectively quantify image reconstruction uncertainty without the need for training data. We parameterize the target posterior using a flow-based model and minimize their Kullback-Leibler (KL) divergence to achieve accurate uncertainty estimation. To bolster stability, we introduce a robust flow-based model with bi-directional regularization and enhance expressivity through gradient boosting. Additionally, we incorporate a space-filling design to achieve substantial variance reduction on both latent prior space and target posterior space. We validate our method on several benchmark tasks and two real-world applications, namely fastMRI and black hole image reconstruction. Our results indicate that our method provides reliable and high-quality image reconstruction with robust uncertainty estimation.
\end{abstract}

\section{Introduction}
In computer vision and image processing, computational image reconstruction is a typical inverse problem where the goal is to learn and recover a hidden image $\mathbf x$ from directly measured data $\mathbf y$ via a forward operator $\mathcal{F}$. Such mapping $\mathbf y = \mathcal{F} (\mathbf x)$, referred to as the \emph{forward process} is often well-established. Unfortunately, the \emph{inverse process} $\mathbf x = \mathcal{F}^{-1} (\mathbf y)$, proceeds in the opposite direction, which is a nontrivial task since it is often ill-posed. A regularized optimization is formulated to recover the hidden image $\mathbf x^*$: 
\begin{equation}
    \mathbf x^* =\argmin_{\mathbf x} \left\{\mathcal{L}(\mathbf y, \mathcal{F}(\mathbf x)) + \lambda~ \omega(\mathbf x) \right\}, \label{eq:eq1}
\end{equation}
where $\mathcal{L}$ is a loss function to measure the difference between the measurement data and the forward prediction, $\omega$ is a regularization function, and $\lambda$ is a regularization weight. The regularization function, including $\ell_1$-norm and total variation, are often used to constrain the image to a unique inverse solution in under-sampled imaging systems \cite{bouman1993generalized,strong2003edge}. 

Recent trends have focused on using deep learning for computational image reconstruction, which does not rely on an explicit forward model or iterative updates but performs learned inversion from representative large datasets \cite{zhu2018image,belthangady2019applications}, with applications in medical science, biology, astronomy and more. However, most of these existing studies in regularized optimization \cite{natterer2001mathematical,park2003super} and feed-forward deep learning approaches \cite{ulyanov2018deep,belthangady2019applications,wang2020deep} mainly focus on pursuing a unique inverse solution by recovering a single point estimate. This leads to a significant limitation when working with underdetermined systems where it is conceivable that multiple inverse image solutions would be equally consistent with the measured data \cite{barbano2020quantifying,sun2020deep}. Practically, in many cases, only partial and limited measurements are available which naturally leads to a \emph{reconstruction uncertainty}. Thus, a reconstruction using a point estimate without uncertainty quantification would potentially mislead the decision-making process \cite{zhang2019reducing,zhou2020bayesian}. Therefore, the ability to characterize and quantify reconstruction uncertainty is of paramount relevance. In principle, Bayesian methods are an attractive route to address the inverse problems with uncertainty estimation. However, in practice, the exact Bayesian treatment of complex problems is usually intractable \cite{zhang2021modern}. The common limitation is to resort to inference and sampling, typically by Markov Chain Monte Carlo (MCMC), which are often prohibitively expensive for imaging problems due to the \emph{curse of dimensionality}. 

For nonlinear, non-convex image reconstruction problems, a deeper architecture with enough expressivity may be required to approximate the complex posterior distribution. However, increasing model depth will result in overfitting, as well as making sampling and computing inefficient \cite{zhang2019learning}. The essential assumption of invertibility is also potentially violated along with instability issues caused by the aggregation of numerical errors. The imprecision and variation in flow-based models induce additional uncertainties. Moreover, the variation in posterior distribution caused by latent space sampling is also non-negligible for the evaluation of the total reconstruction uncertainty. These challenges make accurate uncertainty estimation to be a nontrivial task. 

\vspace{10pt}
\noindent {\bf Main contributions.} 
We present a novel uncertainty-aware method for achieving reliable image reconstruction with an accurate estimation of data uncertainty resulting from measurement noise and sparsity. Our approach leverages a deep variational framework with robust generative flows and variance-reduced sampling to accurately characterize and quantify reconstruction uncertainty without any training data. We propose a flow-based variational approach to approximate the posterior distribution of a target image, minimizing model uncertainties by building a robust flow-based model with enhanced stability through bi-directional regularization and improved flexibility through gradient boosting. We conservatively propagate the statistics of latent distribution to the posterior through a deterministic invertible transformation, replacing simple random sampling with generalized Latin Hypercube Sampling to achieve significant variance reduction on posterior approximation. We demonstrate our method on fastMRI reconstruction and interferometric imaging problems, showing that it achieves reliable and high-quality reconstruction with accurate uncertainty evaluation.

\section{Background}
\noindent \textbf{Normalizing flows.}
Generative models, such as GANs and VAEs, are intractable for explicitly learning the probability density function which plays a fundamental role {\color{black}in} uncertainty estimation. Flow-based generative models overcome this difficulty with the help of \textit{normalizing flows} (NFs), which describe the transformation from a latent density $\mathbf z_0 \sim \pi_0(\mathbf z_0)$ to a target density $\tau (\mathbf x)$, where $ \mathbf x = \mathbf{z}_K \sim \pi_K (\mathbf z_K)$ through a sequence of invertible mappings $\mathcal{T}_k: \mathbb{R}^d \rightarrow \mathbb{R}^d,  k=1,...,K$. By using the change of variables rule 
\begin{equation}
    \tau(\mathbf x) = \pi_k(\mathbf z_k) = \pi_{k-1}(\mathbf z_{k-1}) \left| \det \frac{\partial \mathcal{T}_k^{-1}}{\partial \mathbf z_{k-1}}\right|,  
\label{eq:nf}
\end{equation}
where the target density $\pi_{K}(\mathbf z_K)$ obtained by successively transforming a random variable $\mathbf z_0$ through a chain of $K$ transformations $ \mathbf z_K = \mathcal{T}_K \circ \cdots \circ \mathcal{T}_1 (\mathbf z_0) $ is 
\begin{equation*}
  \log \tau(\mathbf x) = \log{\pi_K}(\mathbf z_K) = \log{\pi_0(\mathbf z_0)} - \sum_{k=1}^K \log{\left| \det \frac{\partial \mathcal{T}_k}{\partial \mathbf z_{k-1}}\right|}
\end{equation*}
where each transformation $\mathcal{T}_k$ must be sufficiently \emph{expressive} while being theoretically \emph{invertible} and efficient to compute the Jacobian determinant. Affine coupling functions \cite{dinh2016density,kingma2018glow} are often used because they are simple and efficient to compute. However, these benefits come at the cost of expressivity and flexibility; many flows must be stacked to learn a complex representation. 

\vspace{10pt}
\noindent \textbf{Density estimation.}
Assuming that samples $\left\{ \bm x_i \right\}_{i=1}^M$ drawn from a probability density $p(\mathbf x)$ are available, our goal is to learn a flow-based model $\tau_{\phi}(\mathbf x)$ parameterized by the vector $\phi$ through a transformation $\mathbf x=\mathcal{T}(\mathbf z)$ of a latent density $\pi_0(\mathbf z)$ with  $\mathcal{T} = \mathcal{T}_K \circ \cdots \circ \mathcal{T}_1$ as a $K$-step flow. This is achieved by minimizing the KL-divergence $\textup{D}_{\textup{KL}} = \textup{KL}(p(\mathbf x) \parallel \tau_{\phi}(\mathbf x))$, which is equivalent to maximum likelihood estimate (MLE).

\vspace{10pt}
\noindent \textbf{Variational inference.}
The goal is to approximate the posterior distribution $p$ through a variational distribution $\pi_K$ encoded by a flow-based model $\tau_{\phi} (\mathbf x)$, which is tractable to compute and draw samples. This is achieved by minimizing the KL-divergence $D_{\textup{KL}} = \textup{KL} (\pi_K \parallel p)$, which is equivalent to maximizing an evidence lower bound 
\begin{equation*}
    \mathcal{V}_{\phi} (\mathbf x) = \mathbb{E}_{\mathbf z_K \sim \pi_K (\mathbf z_K)} \left[-\log p(\mathbf x, \mathbf z_K ) + \log \pi_K (\mathbf z_K | \mathbf x) \right] .
\end{equation*}

\noindent \textbf{Evaluation metrics for deep generative models.}
Designing indicative evaluation metrics for generative models and samples remains a challenge. A commonly used metric for measuring the similarity between real and generated images has been the Fréchet Inception Distance (FID) score \cite{heusel2017gans} but it fails to separate two critical aspects of the quality of generative models: \emph{fidelity} that refers to the degree to which the generated samples resemble the real ones, and \emph{diversity}, which measures whether the generated samples cover the full variability of the real samples. Sajjadi \textit{et al.}~\cite{sajjadi2018assessing} proposed the two-value metrics (\emph{precision} and \emph{recall}) to capture the two characteristics separately. Recently, Naeem \textit{et al.}~\cite{naeem2020reliable} introduced two reliable metrics (\emph{density} and \emph{coverage}) to evaluate the quality of the generated posterior samples and measure the difference between them and ground truth. \emph{density} improves upon the precision metric by dealing with the overestimation issue and \emph{coverage} instead of recall metric is to better measure the diversity by building the nearest neighbor manifolds around the true samples.


\section{Methodology}
\subsection{Deep Variational Framework}
Our goal is to build a deep variational framework to accurately estimate the data uncertainty quantified by an approximation of the posterior distribution. The regularized optimization for solving inverse problems can be written in terms of data fidelity (data fitting loss) and regularity:
\begin{equation}
\begin{aligned}
    \mathbf{x}^* &= \argmin_{\mathbf x} \left\{ \mathcal{L}_{\textup{D}} (\mathbf y, \mathcal F(\mathbf x)) + \lambda \omega(\mathbf x)\right\} 
    \\ &= \argmin_{\mathbf x} \big \{ \underbrace{ \parallel \mathbf y - \mathcal F(\mathbf x) \parallel^2}_{\textup{Data fidelity}} + \underbrace{\lambda \omega(\mathbf x)}_{\textup{Regularity}} \big \}. \label{eq:regu}
\end{aligned}
\end{equation}
Assuming the forward operator $\mathcal{F}$ is known and the measurement noise statistics are given, we can reformulate the inverse problem in a probabilistic way. In the Bayesian perspective, the regularized inverse problem in Eq.~\ref{eq:regu} can be solved by Bayesian inference but aims to maximize the posterior distribution by searching a point estimator $\mathbf x^*$:
\begin{equation}
\begin{aligned}
    \mathbf x^* &= \argmax_{\mathbf x} \big \{ \underbrace{ \log p(\mathbf x | \mathbf y)}_{\textup{Posterior}} \big \} 
    \\ &=  \argmax_{\mathbf x} \big \{ \underbrace{\log p(\mathbf y | \mathbf x)}_{\textup{Data likelihood}} + \underbrace{\log p(\mathbf x )}_{\textup{Prior}} \big\}, \label{eq:bayes}
\end{aligned}
\end{equation}
where the prior distribution $p(\mathbf x)$ (e.g., image prior \cite{ulyanov2018deep} in reconstruction problems) defines a similar regularization term and data likelihood $p(\mathbf y | \mathbf x)$ corresponds to the data fidelity in Eq.~\ref{eq:regu}.  If we parameterize the target $\mathbf x$ using a generative model $\mathbf x = \mathcal{T}_{\bm \phi}(\mathbf z), \mathbf z \sim \mathcal{N}(\bm 0, \bm I)$ with model parameter $\bm \phi$, an approximate posterior distribution $\tau_{\bm \phi^*} (\mathbf x)$ is obtained by minimizing the KL-divergence between the generative distribution and the target posterior distribution  
\begin{equation}
\begin{aligned}
 \bm \phi^* &= \argmin_{\bm \phi} \textup{KL}(\tau_{\bm \phi} (\mathbf x) \parallel p(\mathbf x | \mathbf y ))   
 \\ &= \argmin_{\bm \phi} \mathbb{E}_{\mathbf x \sim \tau_{\bm \phi} (\mathbf x)} [-\log p(\mathbf y | \mathbf x) 
 \\ &- \log p(\mathbf x) + \log \tau_{\bm \phi} (\mathbf x) ]. \label{eq:kl1}
\end{aligned}
\end{equation}
Unfortunately, the probability density (likelihood) $\tau_{\bm \phi} (\mathbf x)$ cannot be exactly evaluated by most existing generative models, such as GANs or VAEs. Flow-based models offer a promising approach to computing the likelihood exactly via the change of variable theorem with invertible architectures. Therefore, we can reformulate the equation in terms of a flow-based model as  
\begin{equation}
\begin{aligned}
 \bm \phi^* &= \argmin_{\bm \phi} \mathbb{E}_{\mathbf z \sim \pi(\mathbf z)} \left[ -\log p(\mathbf y | \mathcal{T}_{\bm \phi} (\mathbf{z})) \right.
 \\ & \left. -\log p(\mathcal{T}_{\bm \phi} (\mathbf z)) +  \log \pi(\mathbf z) -\log \big|\det \nabla_{\mathbf z} \mathcal{T}_{\bm \phi}(\mathbf z) \big| \right].
\end{aligned}
\end{equation}
Replacing data likelihood and prior terms by using data fidelity loss and regularization function in Eq.~\ref{eq:regu}, we can define a new optimization problem where it can be approximated by a Monte Carlo method in practice: 
\begin{equation}
\begin{aligned}
 \bm \phi^* & = \argmin_{\bm \phi} \mathbb{E}_{\mathbf z \sim \pi(\mathbf z)} \left[ \mathcal{L}_{\textup{D}} (\mathbf y, \mathcal F(\mathcal{T}_{\bm \phi}(\mathbf z))) + \lambda \omega(\mathcal{T}_{\bm \phi}(\mathbf z))
 \right. \\ & \left. +\log \pi(\mathbf z) -\log \big|\det \nabla_{\mathbf z} \mathcal{T}_{\bm \phi}(\mathbf z) \big| \right] 
 \\ &= \argmin_{\bm \phi} \sum_{j=1}^{M} \bigg[ \mathcal{L}_{\textup{D}} (\mathbf y, \mathcal F(\mathcal{T}_{\bm \phi}(\mathbf z_j))) + \lambda \omega(\mathcal{T}_{\bm \phi}(\mathbf z_j)) 
 \\ &- \underbrace{\log \big|\det \nabla_{\mathbf z} \mathcal{T}_{\bm \phi}(\mathbf z_j) \big|}_{\textup{Entropy}}  \bigg], \label{eq:total_loss}
\end{aligned}
\end{equation}
where $\pi(\mathbf z)$ is a constant and $\log \big|\det \nabla_{\mathbf z} \mathcal{T}_{\bm \phi}(\mathbf z_j) \big|$ is entropy that is critical to encourage sample diversity and exploration to avoid {\color{black}generative models} from collapsing to a deterministic solution. 

Note that the flow-based model is very critical and sensitive to uncertainty estimation and quantification within this variational framework. To perform accurate data uncertainty estimation, the uncertainty associated with the flow-based model must be minimized. To this end, we develop a Robust Generative Flow (RGF) model with enhanced stability, expressivity, and flexibility, while conserving efficient inference and sampling without increasing architecture depth. 
\begin{figure*}[h!] 
  \begin{minipage}[h]{0.65\textwidth} 
    \centering 
    \vspace{-0.3cm}
    \includegraphics[width=0.95\textwidth]{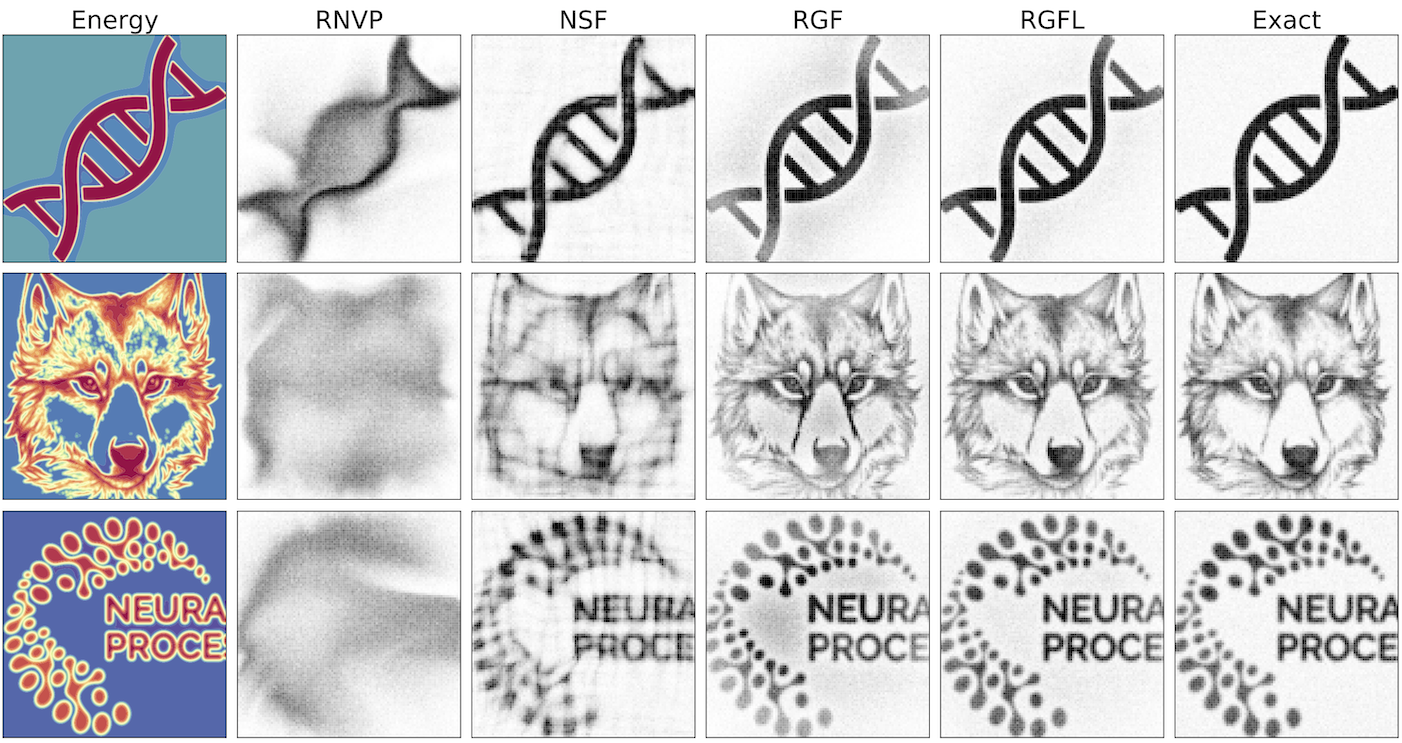} 
    \caption{Sampling of 2D densities using variational inference with energy function.} 
    \label{fig:2d} 
  \end{minipage}%
  \hspace{0.3cm}
  \begin{minipage}[h]{0.3\textwidth}  
  \footnotesize
    \centering
    \begin{tabular}{@{}c|cccc@{}}
    \toprule
    Metrics    & RNVP   & NSF     & RGF     & RGFL    \\ \midrule
    Precision$\uparrow$  & 0.951 & 0.988   & 0.992  & {\bf 0.996}  \\
    Recall$\uparrow$    & 0.996 & 0.994  & {\bf 0.997}  & {\bf 0.997}  \\
    Density$\uparrow$   & 0.823 & 0.962 & 0.987 & {\bf 0.989} \\
    Coverage$\uparrow$ & 0.909 & 0.926  & 0.950    & {\bf 0.961}  \\\midrule
    Metrics    & RNVP   & NSF     & RGF     & RGFL    \\ 
    Precision$\uparrow$ & 0.988 & 0.993  & 0.994  & {\bf 0.997}  \\
    Recall$\uparrow$    & 0.997 & 0.998  & 0.998  & {\bf 0.999}  \\
    Density$\uparrow$   & 0.889 & 0.948  & 0.989  & {\bf 1.001}  \\
    Coverage$\uparrow$ & 0.912 & 0.951  & 0.946  & {\bf 0.964}  \\\midrule
   Metrics    & RNVP   & NSF     & RGF     & RGFL    \\ 
    Precision$\uparrow$ & 0.969 & 0.987  & 0.993  & {\bf 0.994}  \\
    Recall$\uparrow$    & 0.994 & {\bf 0.996}  & {\bf 0.996}  & {\bf 0.996}  \\
    Density$\uparrow$   & 0.829 & 0.942  & 0.987  & {\bf 1.003}  \\
    Coverage$\uparrow$ & 0.877 & 0.934  & 0.940  & {\bf 0.965}  \\ \bottomrule
    \end{tabular}    \makeatletter\def\@captype{table}\makeatother\caption{Evaluation metrics.}
    \label{table:metrics} 
  \end{minipage} 
\end{figure*}

\subsection{Robust Generative Flows (RGF)}
\subsubsection{Stability of invertible architectures}
The proposed variational framework relies on the essential assumption of the theoretical invertibility in flow-based models. However, this assumption is challenged by recent studies \cite{behrmann2019invertible,behrmann2021understanding,kobyzev2020normalizing} with findings that the commonly used invertible architectures $\mathcal{T}_{\bm \phi}$, such as additive and affine coupling blocks, suffer from exploding inverses and thus prone to becoming numerically non-invertible \cite{hoffman2019robust}, which will violate the assumption underlying their main advantages, including efficient sampling and exact likelihood estimation. Typically, the coupling blocks show an exploding inverse effect because the singular values of the forward mapping tend to zero as depth increases. The numerical errors introduced in both forward and inverse mapping will aggravate the imprecision and instability which renders the architecture non-invertible and results in uncontrollable model uncertainties that are intractable to characterize. 

The stability of invertible neural network (INN) architectures can be analyzed by the property of bi-Lipschitz continuity if there exists a constant $L:=\textup{Lip}(\mathcal{T})$ and a constant $L^*:=\textup{Lip}(\mathcal{T}^{-1})$ such that for all $x_1, x_2; y_1, y_2 \in \mathbb{R}^d$
\begin{equation}
\begin{aligned}
    || \mathcal{T}(x_1)- \mathcal{T}(x_2)|| \le L||x_1-x_2||, 
    \\\ || \mathcal{T}^{-1}(y_1)- \mathcal{T}^{-1}(y_2)|| \le L^*||y_1-y_2||.
\end{aligned}
\end{equation}
Penalty terms on the Jacobian can be used to enforce INN stability locally. If $\mathcal{T}$ is Lipschitz continuous and differentiable, we have $\textup{Lip}(\mathcal{T}) = \sup_{\mathbf{x}  \in \mathbb{R}^d} || J_{\mathcal{T}}(\mathbf{x}) ||_2$. Instead of estimating $\textup{Lip}(\mathcal{T})$ using random samples, we use finite differences (FD) to approximate $\textup{Lip}(\mathcal{T})$ as 
\begin{equation}
\begin{aligned}
   \textup{Lip}(\mathcal{T}) &= \sup_{\mathbf{x}  \in \mathbb{R}^d} || J_{\mathcal{T}}(\mathbf{x}) ||_2  \\
   & \approx \sup_{\mathbf{x} \in \mathbb{R}^d} \sup_{||\nu||_2=1} \frac{1}{\varepsilon}|| \mathcal{T}(\mathbf x) - \mathcal{T}(\mathbf x + \varepsilon \nu) ||_2, \label{eq:fd_loss}
\end{aligned}
\end{equation}
where $\varepsilon>0$ is a step size in FD. {\color{black}This penalty term (as a regularizer) can be added to the loss function on both directions $\mathcal{T}$ and $\mathcal{T}^{-1}$ such that we have a stable forward and inverse mapping}. This bi-directional FD regularization can remedy the non-invertible failures in many coupling blocks including spline function.    

{\color{black}
\subsubsection{Enhanced expressivity and flexibility} 
Recent trends in NFs have focused on creating deeper, more complex transformations to increase the flexibility of the learned distribution. However, a deeper structure of flows often renders instability and uncertainty caused by aggregation of the numerical errors. Also, with greater model complexity comes a greater risk of overfitting while slowing down training, sampling, and inference \cite{giaquinto2020gradient}. To address these issues, we propose to use a gradient boosting approach for increasing the expressiveness of the neural spline flow (NSF) model \cite{durkan2019neural}. Our new model is built by iteratively adding new NF components with gradient boosting, where each new NF component is fit to the residual of the previously trained components. A weight is then learned via stochastic gradient descent for each component, which results in a mixture model structure, whose flexibility increases as more components are added. 

\vspace{5pt}
\noindent{\bf Gradient boosting flow model.} A gradient boosting flow model is constructed by successively adding new components, where each new component $t_{K}^{(c)}$ is a K-step normalizing flow that matches the functional gradient of the loss function from the $(c-1)$ previously trained flow components $\mathcal{T}_{k}^{(c-1)}$. Typically, the gradient boost flow model is constructed by a convex combination of fixed and new flow components:
\begin{equation}
    \mathcal{T}_{K}^{(c)}(\mathbf z ~|~ \mathbf x) = (1-\beta_c) \mathcal{T}_{K}^{(c-1)}(\mathbf z ~|~ \mathbf x)  + \beta_c t_K^{(c)}(\mathbf z ~|~ \mathbf x), \label{eq:mix}
\end{equation}
where $\mathbf x$ are the observed data, $\mathbf z$ are the latent variables and $\beta_c$ is the weight to the new component where $\beta_c \in [0,1]$ to sure the mixture model in Eq.~\ref{eq:mix} is a valid probability distribution. 

To pursue a variational posterior that closely matches the true posterior, which corresponds to the reverse KL-divergence $\textup{KL}(\tau_{\bm \phi} (\mathbf x) \parallel p(\mathbf x | \mathbf y ))$. Thus, we seek to minimize the variational bound:
\begin{equation}
    \mathcal{V}_{\phi}(\mathbf x) = \mathbb{E}_{\mathcal{T}_{K}^{(c)}} [\log \mathcal{T}_{K}^{(c)}(\mathbf z_K ~|~ \mathbf x) -\log p(\mathbf y ~|~ \mathbf x, \mathbf z_K) ]. \label{eq:gb_kl_obj}  
\end{equation}

\noindent{\bf Boosting components updating.} Given the objective function in Eq.~\ref{eq:gb_kl_obj}, we proceed with deriving updates for new boosting components. First, at the current stage $c$, we let  $\mathcal{T}_K^{(c-1)}$ to be fixed, and the target is learning the component $t_{K}^{(c)}$ and the weight $\beta_c$ based on functional gradient descent (FGD) \cite{mason1999boosting}. We take the gradient of objective in Eq.~\ref{eq:gb_kl_obj} with respect to $\mathcal{T}_K^{(c)}$ at $\beta_c \rightarrow 0$:
\begin{equation}
    \nabla_{\mathcal{T}_K^{(c)}} \mathcal{V}_{\phi} (\mathbf x) \big|_{\beta_c \rightarrow 0} = -\log \frac{p(\mathbf y ~|~ \mathbf x, \mathbf z_K) }{\mathcal{T}_{K}^{(c-1)} (\mathbf z ~|~ \mathbf{x})}.  \label{eq:gradient} 
 \end{equation}
 Since $\mathcal{T}_K^{(c-1)}(\mathbf z ~|~ \mathbf x)$ are the fixed components, then minimizing the loss $\mathcal{V}_{\phi} (\mathbf x)$ can be achieved by selecting a new component $t_{K}^{(c)}$ that has the maximum inner product with the negative of the gradient \cite{mason1999boosting}. As a result, we can choose a $t_{K}^{(c)} (\mathbf z ~|~ \mathbf x)$ based on:
 \begin{equation}
t_{K}^{(c)} (\mathbf z ~|~ \mathbf x) = \argmin_{t_K \in T_K} \mathop{\mathbb{E}} \limits_{t_{K}(\mathbf z |\mathbf x)} \left[ -\log \frac{p(\mathbf y ~|~ \mathbf x, \mathbf z_K^{(c)}) }{\mathcal{T}_{K}^{(c-1)} (\mathbf z_K^{(c)} ~|~ \mathbf{x})} \right] \label{eq:tk_update}
\end{equation}
where $\mathbf z_{K}^{c}$ denotes a sample transformed by component $c$'s flow.

\vspace{5pt}
\noindent{\bf Components weights updating.} Once the $t_{K}^{(c)} (\mathbf z_K ~|~ \mathbf{x})$ is estimated, the gradient boost flow model needs to determine the corresponding $\beta_c \in [0,1]$. However, jointly optimizing both $t_{K}^{(c)}$ and the weights $\beta_c$ is a challenging optimization problem. We therefore consider a two-step optimization strategy which means that we first train $t_{K}^{(c)}$ until convergence and then optimize the corresponding weight $\beta_c$ by using the objective in Eq.~\ref{eq:tk_update}.  Similarly, the weights on each component can be updated by using the gradient of the loss $\mathcal{V}_{\phi}(\mathbf x)$ with respect to $\beta_c$, as shown in Algorithm \ref{algo:1}. 
\begin{equation}
\begin{aligned}
    \frac{\partial \mathcal{V}_{\phi}(\mathbf x)}{\partial \beta_c} &= \sum_{i=1}^n \left(\mathop{\mathbb{E}} \limits_{t_{K}^{(c)}(\mathbf z |\mathbf x_i)} \left[ \xi_{\beta_c}^{(s-1)}(\mathbf z ~|~ \mathbf x_i) \right] 
    \right. \\& \left. - \mathop{\mathbb{E}} \limits_{\mathcal{T}_{K}^{(c-1)}(\mathbf z |\mathbf x_i)} \left[ \xi_{\beta_c}^{(s-1)}(\mathbf z ~|~ \mathbf x_i) \right] \right)  \label{eq:weight_update}
\end{aligned}
\end{equation}
where $\xi_{\beta_c}^{(s-1)}(\mathbf z ~|~ \mathbf x_i)$ is defined as $\xi_{\beta_c}^{(s-1)}(\mathbf z ~|~ \mathbf x_i) = \log {\mathcal{T}_{K}^{(c-1)} (\mathbf z ~|~ \mathbf{x})} - \log {p(\mathbf y ~|~ \mathbf x, \mathbf z_K) }$.
\begin{algorithm}[h!]
{\color{black}
\caption{\hspace{-0.1cm} Updating component weights $\beta_c$ by SGD} 
\begin{algorithmic}[1] \label{algo:1}
\STATE Requirement and initialization: predefined $C$, step size $\lambda$, tolerance $\epsilon$. Set $s = 0$, $\beta_c^{(0)} = 1/C$
\STATE {\bfseries While} $|\beta_c^{(s)} - \beta_c^{(s-1)}| < \epsilon$ {\bfseries do}
\STATE Generate samples $\mathbf z_{K,i}^{(c-1)} \sim \mathcal{T}_{K}^{(c-1)}(\mathbf z ~|~ \mathbf x_i)$ and $\mathbf z_{K,i}^{(c)} \sim t_K^{(c)}(\mathbf z ~|~ x_i)$ for $i=1,...,n$ \\
\STATE Estimate gradients using the MC method 
$\nabla_{\beta_c} \mathcal{V}_{\psi}(\mathbf x) = 1/n \sum_{i=1}^n \left[\xi_{\beta_c}^{(s-1)}(\mathbf z_{K,i}^{(c)} ~|~ \mathbf x_i) - \xi_{\beta_c}^{(s-1)}(\mathbf z_{K,i}^{(c-1)} ~|~ \mathbf x_i) \right]$
\STATE Update weights $\beta_c^{(s)} = \beta_c^{(s-1)} - \lambda \nabla_{\beta_c}$ and clip weights $\beta_c^{(s)}$ to $[0,1]$
\STATE $s = s + 1$
\RETURN $\beta_c^{(s)}$
\end{algorithmic}}
\end{algorithm}
}
\subsection{Variance-reduced Latent Sampling}
\label{sec:vr}

\begin{figure*}[h!]
    \centering
    \vspace{-0.2cm}
    \includegraphics[width=0.8\textwidth]{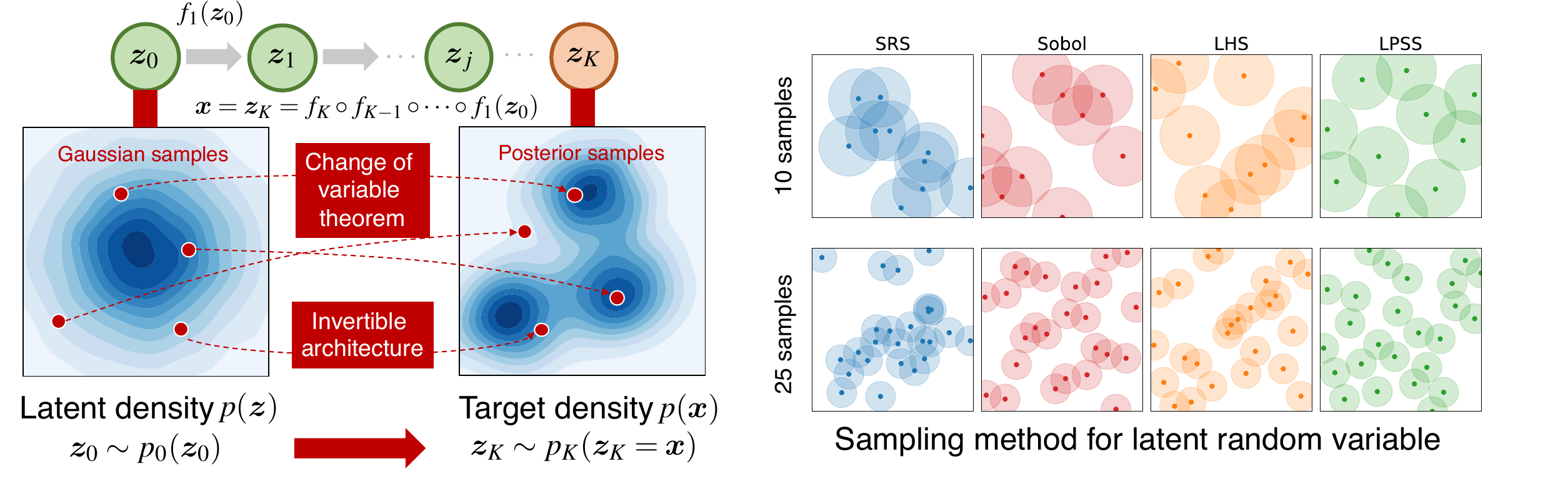}
    \caption{(Left) Illustration of mechanism for normalizing flows. The prior Gaussian samples drawn from the latent distribution {\color{black} are} transformed to the posterior samples that match the target density by using a sequence of invertible mappings. (Right) Illustration of prior samples using four sampling algorithms for latent variables.}
    \label{fig:sampling}
    \vspace{-0.3cm}
\end{figure*}

Rather than using simple random sampling (SRS) with a larger variance, Latin Hypercube Sampling (LHS) is an ideal candidate with variance reduction, which is generalized in terms of a spectrum of stratified sampling, referred to as partially stratified sampling (PSS), which shows to reduce variance associated with variable interaction but LHS reduces variance associated with additive (main) effects. A hybrid combination of LHS and PSS, named Latinized partially stratified sampling (LPSS) proposed by \cite{shields2016generalization} can reduce variance associated with variable interaction and additive effects simultaneously. However, classical LHS used in LPSS suffers from a lack of exploratory capability due to its random pair scheme. To better explore all possible inverse solutions, we propose to leverage maximin criteria - maximizes the minimum distance between all pairs of points, 
\begin{equation}
    X_n = \argmax \min\left\{d(\mathbf x_i, \mathbf x_k): i \neq k =1,...,M \right\}  \label{eq:lhs}
\end{equation}
where $d(\mathbf x, \mathbf x^{\prime}) = \sum_{k=1}^M(\mathbf x_i - \mathbf x_k^{\prime})^2$ is the Euclidean distance. Instead of random pair, these criteria will greatly improve the space-filling properties of LPSS, specifically in high-dimensional space. Note that, unlike quasi-Monte Carlo (QMC) methods \cite{caflisch1998monte}, e.g., Sobol sequence shown in Fig. \ref{fig:sampling} as well, which are limited in high dimensional problem \cite{kucherenko2015exploring}, but LPSS performs well with space-filling property and variance reduction in high dimensions.   


Algorithm \ref{algo:2} shows the workflow of our proposed variational framework with robust generative flows and variance-reduced latent sampling given a single measurement data. 
\vspace{-0.1cm}
\begin{algorithm}[h!]
\caption{\hspace{-0.1cm} RGF in variational perspective}
\begin{algorithmic}[1] \label{algo:2}
\STATE{\bf Requirements}: RGF model $\mathcal{T}_{\bm \phi}$ parameterized by ${\bm \phi}$, number of gradient boosting (GB) component $C$, FD regularization coefficient $\lambda$, forward operator $\mathcal{F}$, measurement data $\mathbf{y}$, training batch size $b_z$, evaluation batch size $e_z$, 
\STATE Draw random samples from the latent space $\mathbf{z}_j \sim p(\mathbf z)$ using LPSS, where the sample size is $b_z \times i_z$ 
\STATE Generate image samples by $\mathbf{x}_j = \mathcal{T}_{\bm \phi}(\mathbf z_j)$ where $\mathcal{T}_{\bm \phi}(\mathbf z_j)$ is defined by initial gradient boosting (GB) mixture model in Eq.~\ref{eq:mix} with bi-directional FD regularized loss in Eq.~\ref{eq:fd_loss}. 
\STATE Predict the measurement $\mathbf y$ by evaluating the forward operator $\mathbf{y}_j = \mathcal{F}(\mathbf{x}_j) = \mathcal{F}(\mathcal{T}_{\bm \phi}(\mathbf z_j)) $  
\STATE Evaluate the total loss $\mathcal{L}_{\textup{total}}$ defined in Eq.~\ref{eq:total_loss} by a summation of the data fidelity, prior and entropy loss. 
\STATE Update new gradient boosting components based on  Eq.~\ref{eq:tk_update} with a two-stage optimization strategy 
\STATE Update boosting component weights according to Eq.~\ref{eq:weight_update} with stochastic gradient descent in Algorithm 1. 
\STATE After training is done, generate posterior samples, and estimate the statistical quantities. 
\end{algorithmic}
\end{algorithm}
\vspace{-0.4cm}
\subsection{RGF Representational Capability}
Next, we demonstrate that our proposed RGF {\color{black} can} improve the representational capability of deterministic NFs at a given network size. Here we use images to define complicated two-dimensional distributions as the target distributions to be sampled. This benchmark inspired by \cite{wu2020stochastic} aims at generating high-quality images from the exact density to compare the performance of different generative models with four blocks. Fig.\ref{fig:2d} shows the sampling of DNA, Fox, and NeurIPS logo cases with different methods.  As expected, RealNVP \cite{dinh2016density} has limitations in representational capability, which results in a blurred image without detailed structures. NSF \cite{durkan2019neural} outperforms RNVP with a better structure on 2D images, but still fails to resolve details, specifically in the Fox and NeurIPS logo cases, at the selected neural architecture. Note that all the flow-based models tend to have a better representational capability as the depth increases but we fix their depth in four blocks to perform a fair comparison. 

Our proposed RGF (with SRS) achieves high-quality approximation through a combination of gradient boosting and stable regularization with a shallow architecture. Although the samples in RGF are close to the ground truth, their performance can be further improved by using LPSS. Although visual differences may be slight between RGF and RGFL, we differ them with a quantitative comparison using evaluation metrics (see Table \ref{table:metrics}). RGFL shows superior performance on both fidelity and diversity aspects of the generated images. In other words, the RGF model boosted by LPSS achieves not only a more accurate approximation to the true samples but also better captures the uncertainty (variation) in real samples. Both advantages are useful for quantifying the reconstruction uncertainty.
\section{Experiments}
\subsection{Experimental setup}
\noindent \textbf{Tasks and datasets.} 
We evaluate our method on two image reconstruction tasks: (1) compressed sensing MRI, from fastMRI dataset \cite{zbontar2018fastmri} with images of size $320\times320$; and (2) Interferometric imaging, using blackhole images from the Event Horizon Telescope (EHT) \cite{akiyama2019first1,akiyama2019first2}, with images of size $160\times160$. For both cases, we only use {\bf one data}, which contains {\bf one} specific measurement and the corresponding ground truth. The original images are resized to $128\times128$ for fastMRI and $32\times32$ for blackhole cases.

\vspace{5pt}
\noindent \textbf{Implementation.} We use 6 flow steps and 4 residual blocks for each step so that we have a total of 24 transformations in our RGF model. The whole model is trained on a single V100 GPU by using 20000 epochs with a batch size of 32 and an initial learning rate of 1E-4 in Adam. 

\vspace{5pt}
\noindent \textbf{Evaluation metrics.} After training, 1000 samples drawn from the learned posterior distribution are used to evaluate the \textit{mean}, \textit{standard deviation}, and \textit{absolute error}, which represents the bias between the mean image and ground truth. To evaluate the performance of the generative model, we provide a quantitative evaluation of the sample fidelity using \textit{precision} and \textit{density}, and the sample diversity using \textit{recall} and \textit{coverage}.

\vspace{5pt}
\noindent \textbf{Baselines.} We compare our proposed methods with a couple of methods including (1) conditional variational autoencoder (cVAE) \cite{sohn2015learning}, which is typically used as a baseline method; (2)
Deep Probabilistic Imaging (DPI), proposed by \cite{sun2020deep}, mainly uses the RealNVP model for image Reconstruction; (3) GlowIP, which uses invertible generative models for inverse problems and image reconstruction \cite{asim2020invertible}; (4) NSF, which is neural spline flows \cite{durkan2019neural} used in our deep variational framework. RGFL is our proposed method which uses robust generative flows (RGF) with Latinized partially stratified sampling (LPSS). Although some recent methods \cite{wang2020deep,whang2021composing,whang2021solving}, look very promising, we can not compare them if there are no available open-source codes. 
\begin{figure}[!h]
    \centering
    \vspace{-0.3cm}
    \includegraphics[width=0.47\textwidth]{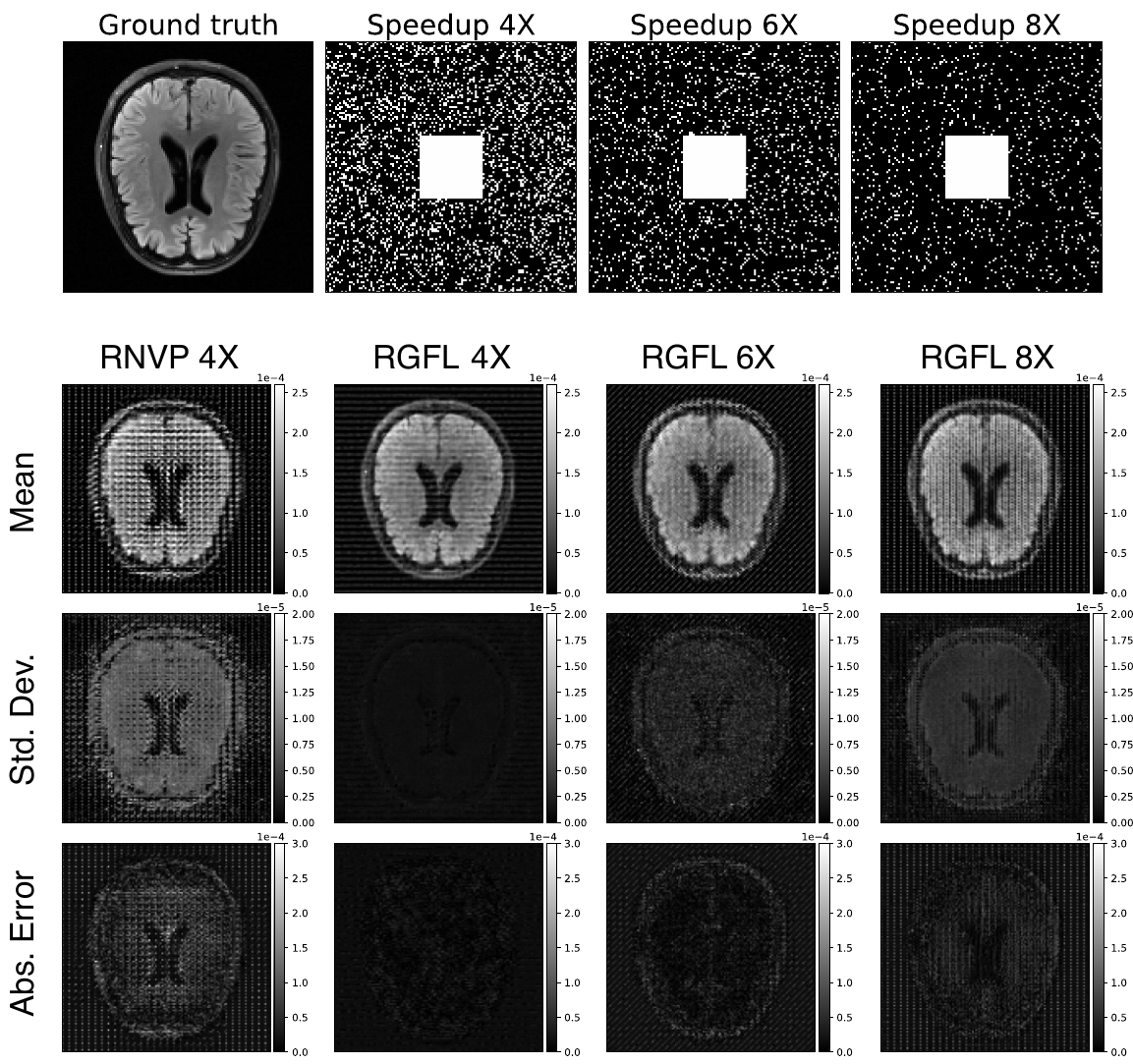}
    \includegraphics[width=0.47\textwidth]{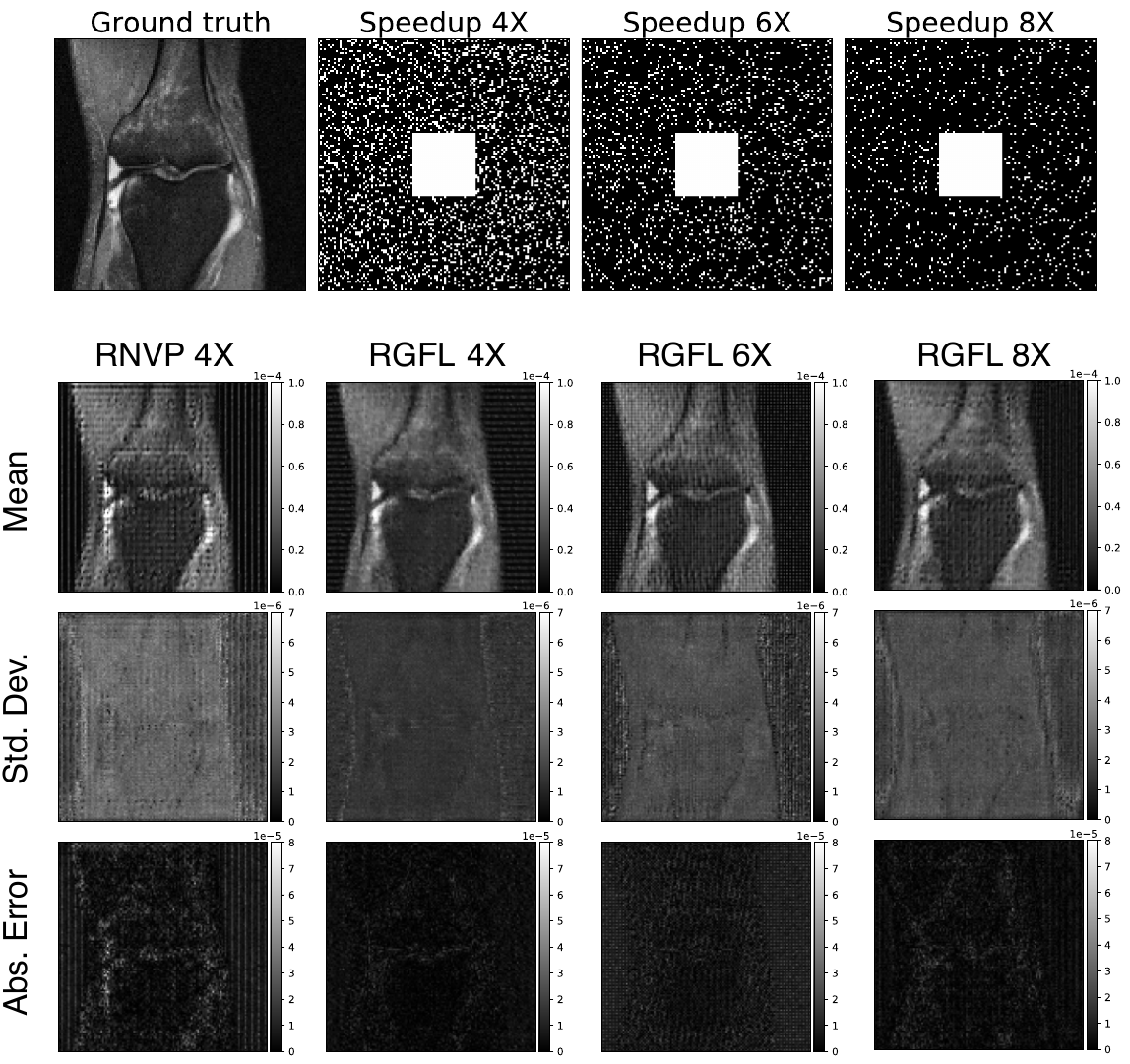}
    \caption{FastMRI reconstruction of brain case (up) and knee case (bottom) at three different acceleration speedup factors: 4X, 6X, and 8X (each shown in a column). Row 1 shows the ground truth and sampling masks for each case. Row 2-4 shows the mean, standard deviation, and absolute error for the estimated posterior samples. }
    \label{fig:mri_brain}
        \vspace{-10pt}
\end{figure}

\begin{table}[h!]
\footnotesize
\centering
\begin{tabular}{@{}c|cccccccccccccccc@{}}
\toprule
Metrics                 & \multicolumn{5}{c}{Brain case with 4$\times$ speedup}                                                \\ \midrule
                        & cVAE    & DPI     & NSF       & GrowIP                       & RGFL                           \\
Std. Dev. $\downarrow$  & 8.87E-5 & 1.21E-5 & 8.82E-6   & 6.90E-6                      & {\bf 9.73E-7} \\
Abs. Error $\downarrow$ & 8.72E-5 & 3.31E-5 & 6.91E-6   & 1.83E-6                      & {\bf 1.78E-6} \\
\midrule
                        & \multicolumn{5}{c}{Brain case with 6$\times$ speedup}                                                \\ \midrule
                        & cVAE    & DPI     & NSF       & GrowIP                       & RGFL                           \\
Std. Dev. $\downarrow$  & 1.02E-4 & 1.97E-5 & 1.02E-5   & 7.33E-6                      & {\bf 2.35E-6} \\
Abs. Error $\downarrow$ & 1.39E-4 & 5.70E-5 & 8.04E-6   & 6.99E-6                      & {\bf 4.58E-6} \\
\midrule
                        & \multicolumn{5}{c}{Brain case with 8$\times$ speedup}                                                \\ \midrule
                        & cVAE    & DPI     & NSF       & GrowIP                       & RGFL                           \\
Std. Dev. $\downarrow$  & 2.58E-4 & 3.83E-5 & 2.27E-5   & 9.14E-6                      & {\bf 4.10E-6} \\
Abs. Error $\downarrow$ & 4.65E-4 & 8.81E-5 & 1.33E-5   & 8.04E-6                      & {\bf 7.12E-6} \\
\bottomrule
\end{tabular}
\vspace{-2mm}
\caption{FastMRI brain case results compared to baseline methods.}
\label{tab:brain}
\end{table}

\begin{table}[h!]
\footnotesize
\centering
\begin{tabular}{@{}c|cccccccccccccccc@{}}
\toprule
Metrics                 & \multicolumn{5}{c}{Knee case with 4$\times$ speedup}                                                \\ \midrule
                        & cVAE    & DPI     & NSF       & GrowIP                       & RGFL                           \\
Std. Dev. $\downarrow$  & 7.98E-6 & 5.94E-6 & 4.33E-6                      & 3.84E-6                      & {\bf 1.24E-6} \\ 
Abs. Error $\downarrow$ & 7.99E-6 & 4.83E-6 & 2.55E-6                      & 9.36E-7                      & {\bf 7.03E-7} \\
\midrule
                        & \multicolumn{5}{c}{Knee case with 6$\times$ speedup}                                                \\ \midrule
                        & cVAE    & DPI     & NSF       & GrowIP                       & RGFL                           \\
Std. Dev. $\downarrow$  & 9.53E-6 & 9.33E-6 & 6.48E-6                      & 4.71E-6                      & {\bf 2.05E-6} \\
Abs. Error $\downarrow$ & 1.02E-5 & 6.02E-6 & 4.87E-6                      & 1.96E-6                      & {\bf 9.11E-7} \\
\midrule
                        & \multicolumn{5}{c}{Knee case with 8$\times$ speedup}                                                \\ \midrule
                        & cVAE    & DPI     & NSF       & GrowIP                       & RGFL                           \\
Std. Dev. $\downarrow$  & 3.61E-5 & 1.60E-5 & 9.96E-6                      & 5.18E-6                      & {\bf 3.71E-6} \\
Abs. Error $\downarrow$ & 3.45E-5 & 7.94E-6 & 6.07E-6                      & 2.06E-6                      & {\bf 1.15E-6} \\
\bottomrule
\end{tabular}
\vspace{-2mm}
\caption{FastMRI knee case results compared to baseline methods.}
\label{tab:knee}
\end{table}

\begin{figure*}[h!]
    \centering
    \vspace{-0.4cm}
    \includegraphics[width=0.48\textwidth]{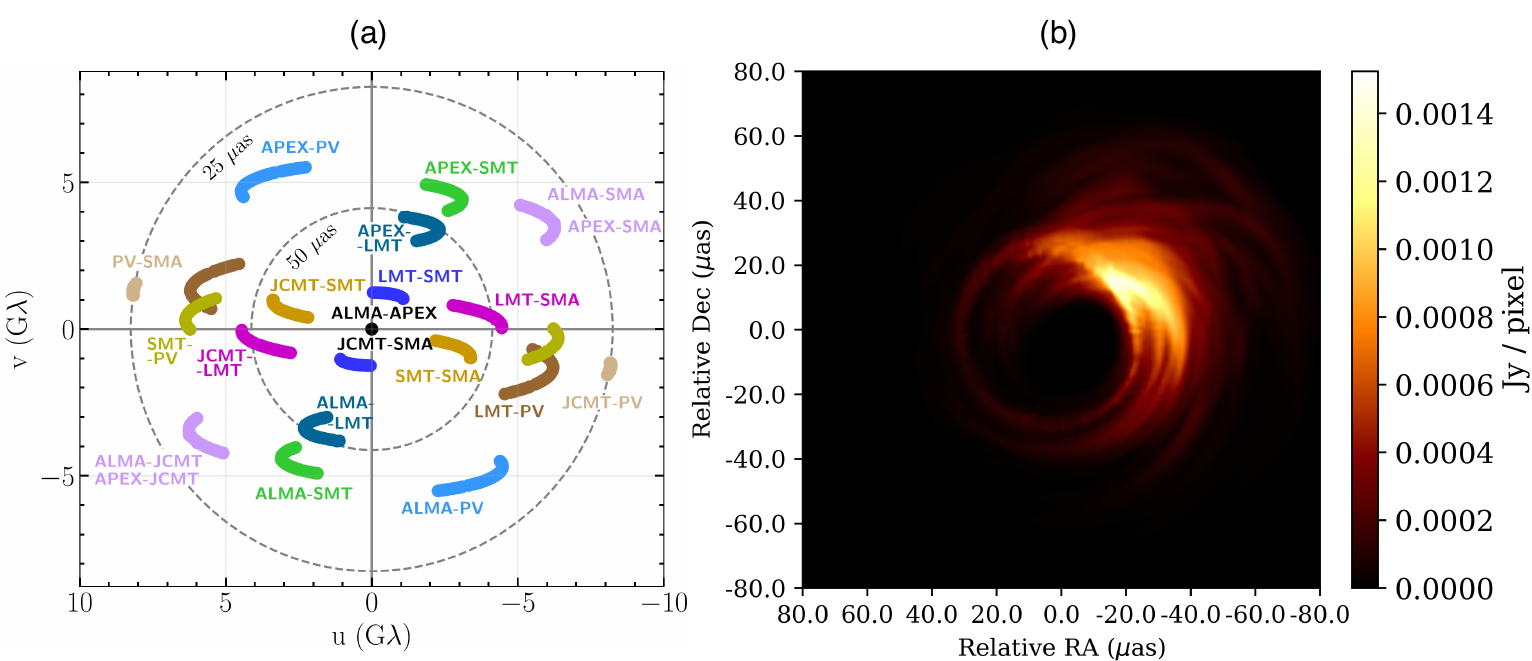}
    \hspace{0.3cm}
    \includegraphics[width=0.48\textwidth]{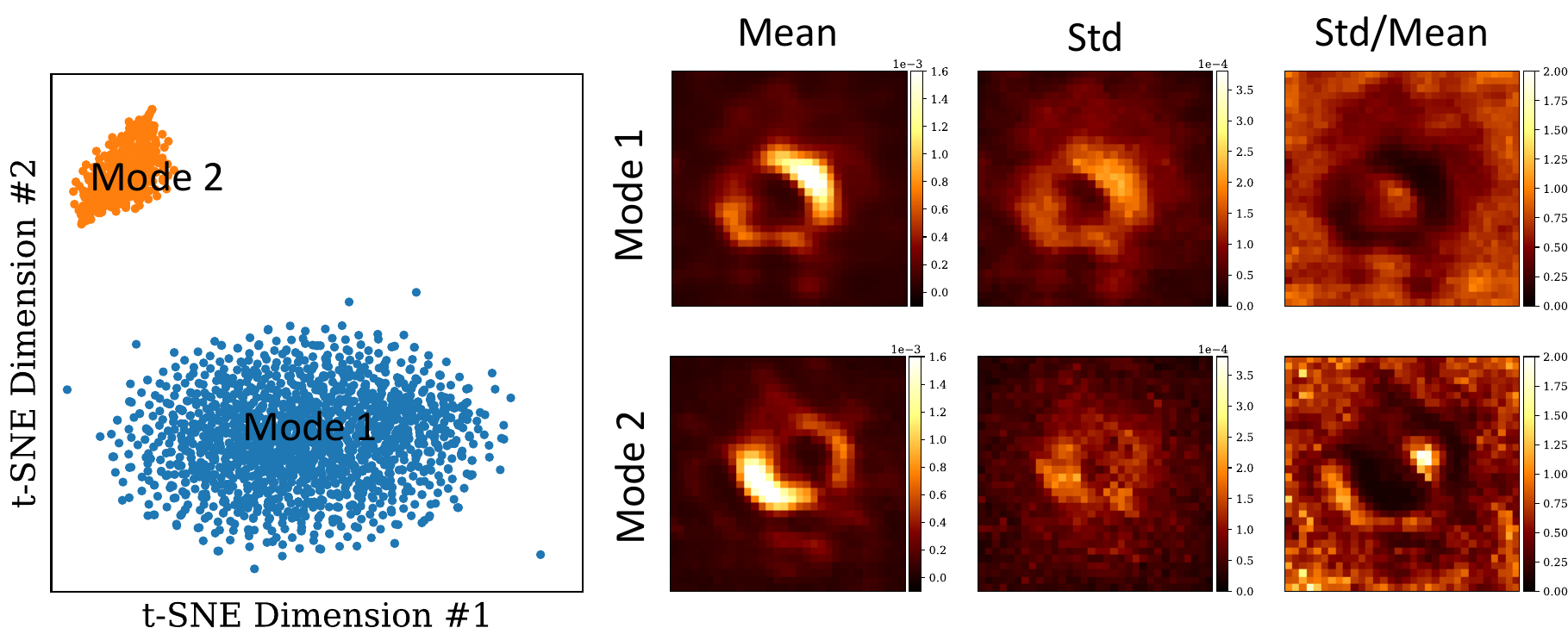}
    \caption{(Left) Measured frequency samples for EHT observing the M87$^*$  \cite{akiyama2019first1,akiyama2019first2} and target synthetic black hole image; (Right) t-SNE result of posterior samples and statistics of the posterior samples on each model  \cite{sun2020deep}.}
    \label{fig:black}
    \vspace{-10pt}
\end{figure*}

\subsection{FastMRI Case Study}
Partial and under-sampled noisy measurements in MRI will lead to reconstruction uncertainty. We show that our RGFL method can be successfully applied to quantify the reconstruction variation and bias on two cases (brain and knee) with acceleration factors 4$\times$, 6$\times$, and 8$\times$. Fig. \ref{fig:mri_brain}  shows the reconstruction results with pixel-wise statistics of the estimated posterior distribution. For both cases with a speedup 4$\times$ factor, our RGFL shows a more accurate mean estimate with a smaller absolute error than the DPI baseline with a larger architecture. Our advantage in terms of standard deviation is more significant due to higher model robustness and variance reduction. Although the pixel-wise variance of the reconstruction tends to be larger as the speedup factor increases, our method provides a more reliable image reconstruction compared with other baselines. 

We further use the mean of the standard deviation and the absolute error to quantitatively compare the pixel-wise statistics (see Table \ref{tab:brain} and \ref{tab:knee}). Our method outperforms the other baselines in terms of accuracy (bias) and variation of the reconstruction. Specifically, our estimation achieves significant variance reduction with 1-2 orders of magnitude. Regarding the sample fidelity and diversity metrics, our method also shows competitive performance in most cases.  
\subsection{Interferometric Imaging Case Study} 
Our approach can be also applied to study radio interferometric astronomical imaging which was used to take the first black hole images. Sparse spatial frequency measurements are used to recover the underlying astronomical image (see Fig. \ref{fig:black}). Relatively large telescope-based gain and phase error in the measurement noise leads to a non-convex image reconstruction problem where a challenge is the potential for multi-modal posterior distribution --- different solutions fit the same measurement data visually and reasonably well. In this case, a synthetic black hole is used to illustrate our capability on reconstructed uncertainty estimation and multiple modes detection \cite{sun2020deep}.
\begin{table}[h!]
\footnotesize
\centering
\begin{tabular}{@{}c|cccccc@{}}
\toprule
Metrics & \multicolumn{5}{c}{Reconstructed black hole images (mode 1)}    \\ \midrule
          & cVAE     & DPI    & NSF      & GrowIP  & {\bf RGFL}  \\
Std. Dev.  $\downarrow$   & 1.23E-3    & 6.09E-4 & 5.43E-4 & 2.17E-4 & {\bf 1.20E-4}         \\
Abs. Error $\downarrow$   & 9.76E-4     & 3.85E-4 & 2.65E-4 & {\bf 1.06E-4} & {1.08E-4}            \\ 
\bottomrule
\end{tabular}
\vspace{-2mm}
\caption{Statistical comparison of the estimated posteriors on the black hole image reconstruction (mode 1).}
\label{tab:blackhole}
\end{table}

Fig. \ref{fig:black} shows multiple samples drawn from the learned generative model. Note that two different modes of reconstruction are captured by our RGFL model. t-SNE plots present a clear clustering of samples into two modes. Mode 1 fits the target image better than mode 2 which exhibits roughly 180-degree rotations. We perform statistical analysis for each mode to quantify the mean and standard deviation of the multi-modal posterior. Note that the uncertainty characterized by standard deviation shows a similar shape to the mean estimator. Mode 1, which is close to the correct solution, results in a smaller bias estimator than mode 2 (see Table \ref{tab:blackhole}). In this case, the space-filling sampling would be more important to capture the multi-modal posterior distribution. This can also be observed by the recall and coverage metrics in Table \ref{tab:blackhole}. Our method outperforms the other baselines with more accurate and reliable results.

\section{Related work}
\noindent \textbf{Bayesian deep learning.} Deep learning for solving inverse problems requires uncertainty estimation to be reliable in real settings. Bayesian deep learning \cite{kendall2017uncertainties,khan2018fast,wilson2020bayesian}, specifically Bayesian neural networks \cite{hernandez2015probabilistic} can achieve this goal while offering a computationally tractable way for recovering reconstruction uncertainty. However, exact inference in the BNN framework is not a trivial task, so several variational approximation approaches are proposed to deal with the scalability challenges. Monte Carlo dropout \cite{gal2016dropout} can be seen as a promising alternative approach that is easy to implement and evaluate. Deep ensemble \cite{lakshminarayanan2017simple} methods proposed by combining multiple deep models from different initializations have outperformed BNN. Recent methods on deterministic uncertainty quantification \cite{van2020uncertainty,van2021improving} use a single forward pass but scale well to large datasets. Although these approaches show impressive performance, they rely on supervised learning with paired input-output datasets and only characterize the uncertainty conditioned on a training set. 

\vspace{5pt}
\noindent \textbf{Variational approaches.} Variational methods offer a more efficient alternative approximating true but intractable posterior distribution by an optimally selected tractable distribution family \cite{blei2017variational}. However, the restriction to limited distribution families fails if the true posterior is too complex. Recent advances in conditional generative models, such as conditional GANs (cGANs) \cite{wang2018high}, overcome this restriction in principle, but have limitations in satisfactory diversity in practice. Another commonly adopted option is conditional VAEs (cVAEs) \cite{sohn2015learning}, which outperform cGANs in some cases, but in fact, the direct application of both conditional generative models in computational imaging is challenging because a large number of data is typically required \cite{tonolini2020variational}. This introduces additional difficulties if our observations and measurements are expensive to collect. 

\vspace{5pt}
\noindent \textbf{Deep flow-based models.} Many very recent efforts have been made to solve inverse problems via deep generative models \cite{asim2020invertible,whang2021composing,bora2017compressed,daras2021intermediate,sun2020deep,bi2023accelerating}. The flow-based model is a potential alternative via learning of a nonlinear transformation between the true posterior distribution and a simple prior distribution \cite{rezende2015variational,dinh2016density,kingma2018glow,grathwohl2018ffjord,wu2020stochastic,nielsen2020survae}. These flow-based models possess critical properties: (a) the neural architecture is \emph{invertible}, (b) the forward and inverse mapping is efficiently \emph{computable} and (c) the Jacobian is \emph{tractable}, which allows explicit computation of posterior probabilities. Fully invertible neural networks (INNs) are a natural choice to satisfy these properties and can be built using coupling layers, as introduced in the RealNVP \cite{dinh2016density} which is simple and efficient to compute. Although in principle, RealNVP layers are theoretically invertible, the actual computational of their inverse is not stable and sometimes even non-invertible due to aggravating numerical errors \cite{behrmann2021understanding}. Many efforts have been spent to improve the stability, invertibility, flexibility, and expressivity of the flow-based models \cite{kong2020expressive,nielsen2020survae,wu2020stochastic, wang2023autonf}, which inspired us to extend for the task of computing posterior in real-world imaging reconstruction. 

Recently, Asim \textit{et al.}~\cite{asim2020invertible} {\color{black}focused} on producing a point estimate motivated by the MAP formulation and Wang \textit{et al.}~\cite{whang2021composing} aims at studying the full distributional recovery via variational inference. A follow-up study from \cite{whang2021solving} is to study image inverse problems with a normalizing flow prior, which is achieved by proposing a formulation that views the solution as the maximum a posteriori estimate of the image conditioned on the measurements. Our work is motivated by these recent advances but focuses on how to assess the image reconstruction (data) uncertainty with an explicitly known forward model with very sparse observations. 

\section{Conclusion}
We propose an uncertainty-aware framework that leverages a deep variational approach with robust generative flows and variance-reduced sampling to perform an accurate estimation of image reconstruction uncertainty. The results on multiple benchmarks and real-world tasks demonstrate our advantages in uncertainty estimation. Although the current RGFL methods show superior performance on these imaging reconstruction tasks, the total computational cost will be a major concern if we plan to scale to complex high-resolution images, e.g., large-scale scientific simulations \cite{lavin2021simulation}.

{\small
\bibliographystyle{ieee_fullname}
\bibliography{egbib}
}

\end{document}